\documentclass[letterpaper, 10 pt, conference]{ieeeconf}  

\IEEEoverridecommandlockouts                              

\usepackage{cite}
\usepackage{amsmath,amssymb,amsfonts}
\usepackage{algorithmic}
\usepackage{graphicx}
\usepackage{float}
\usepackage{textcomp}
\usepackage[usenames,dvipsnames,svgnames,table]{xcolor}
\usepackage{array}
\usepackage{gensymb}
\usepackage[utf8]{inputenc}
\usepackage[font=scriptsize]{caption}
\usepackage{subcaption}
\usepackage{comment}
\usepackage{balance}
\usepackage{units}
\usepackage[colorlinks=true,pdfpagemode=UseNone,citecolor=black,linkcolor=black,urlcolor=BrickRed]{hyperref}
\usepackage{cleveref}

\def\BibTeX{{\rm B\kern-.05em{\sc i\kern-.025em b}\kern-.08em
    T\kern-.1667em\lower.7ex\hbox{E}\kern-.125emX}}
    
\usepackage{framed}
\definecolor{shadecolor}{RGB}{180,180,180}

\newcommand{\real}{\mathbb R}

\makeatletter
\renewcommand{\maketag@@@}[1]{\hbox{\m@th\normalsize\normalfont#1}}%
\makeatother
\title{\LARGE \bf Terrain-Adaptive, ALIP-Based Bipedal Locomotion Controller via Model Predictive Control and Virtual Constraints \\

}

\author{Grant Gibson, Oluwami Dosunmu-Ogunbi, Yukai Gong, and Jessy Grizzle
\thanks{The authors are with the College of Engineering and the Robotics Department, University of Michigan, Ann Arbor, MI 48109 USA {\tt\small \{grantgib,wami,ykgong,grizzle\}}@umich.edu. }
}

\begin{document}

\maketitle
\thispagestyle{empty}
\pagestyle{empty}


\setlength{\textfloatsep}{10pt plus 1.0pt minus 2.0pt}
\setlength{\floatsep}{10pt plus 1.0pt minus 2.0pt}
\setlength{\intextsep}{10pt plus 1.0pt minus 2.0pt}
\setlength{\dbltextfloatsep}{5pt plus 1.0pt minus 2.0pt}

\begin{abstract}
This paper presents a gait controller for bipedal robots to achieve highly agile walking over various terrains given local slope and friction cone information. Without these considerations, untimely impacts can cause a robot to trip and inadequate tangential reaction forces at the stance foot can cause slippages. We address these challenges by combining, in a novel manner, a model based on an Angular Momentum Linear Inverted Pendulum (ALIP) and a Model Predictive Control (MPC) foot placement planner that is executed by the method of virtual constraints. The process starts with abstracting from the full dynamics of a Cassie 3D bipedal robot, an exact low-dimensional representation of its center of mass dynamics, parameterized by angular momentum. Under a piecewise planar terrain assumption and the elimination of terms for the angular momentum about the robot's center of mass, the centroidal dynamics about the contact point become linear and have dimension four. Importantly, we include the intra-step dynamics at uniformly-spaced intervals in the MPC formulation so that realistic workspace constraints on the robot's evolution can be imposed from step-to-step. The output of the low-dimensional MPC controller is directly implemented on a high-dimensional Cassie robot through the method of virtual constraints. In experiments, we validate the performance of our control strategy for the robot on a variety of surfaces with varied inclinations and textures.
\end{abstract}

\section{Introduction} \label{Sec:Intro}
\begin{figure}[t]
    \centering
    \includegraphics[width=0.8\columnwidth]{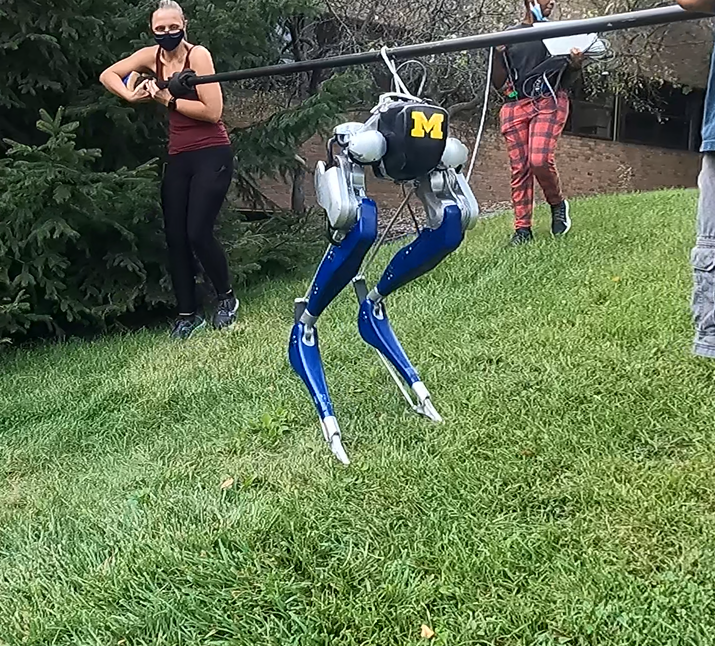}
    \caption{Cassie Blue using a 3D-ALIP inspired MPC gait controller to walk sideways up a 22$^o$ incline on wet grass. Lateral walking with Cassie is much more difficult than longitudinal walking due to tight workspace constraints, which are accounted for in our formulation. }
    \label{fig:cassie_intro}
\end{figure}

This paper contributes to the growing literature on terrain-adaptive locomotion. Our objective is to design a gait (locomotion) controller that enables an agile bipedal robot, such as Cassie in Fig.~\ref{fig:cassie_intro},  to traverse terrain as close to a planned velocity as the physical limits of the robot and terrain conditions allow. We assume that the robot is (a) provided a local planar approximation of the terrain, (b) a local friction cone, and (c) a vector field of desired velocity (speed, heading, and yaw rate) as a function of the robot's current pose and velocity. The integral curves of the vector field provide a family of paths that the robot may follow to reach a goal that is unknown to the local gait controller. These parameters may come from a reactive planner, as in \cite{arslan2019sensor,paternain2017navigation}, or through a human operator and a Radio Control (RC) transmitter, as is done in this paper. 

We make a key simplifying assumption on the terrain, namely, that over robot step-length distances, it can be piecewise approximated by planes, with allowed jumps at the boundaries. This admittedly vague assumption will be made more precise in Sect.~\ref{sec:3DModels}, where we model the center of mass dynamics of the robot. The MPC foot-placement controller plans $N$ robot-steps ahead with a terminal cost that assumes the terrain slope and friction cone remain the same beyond the planning horizon. The ability to include workspace constraints (e.g., avoid self collisions) and an approximate friction cone (e.g., avoid overconfidence on horizontal ground reaction forces) leads to significantly enhanced agility over the original ALIP-based controller introduced in \cite{gong2021one}. In addition, the inclusion of a piecewise linear approximation of the local terrain in the ALIP model enlarges the situations where the approximate zero dynamics analysis and associated stability guarantees in \cite{yukai2022arxiv} are applicable.

\subsection{Related Work} 

\noindent \textbf{Switching Control Based on One-step Ahead Terrain Profile:} Terrain-adaptive locomotion of a simulated 3D humanoid is achieved in \cite{wu2010terrain}. First, in an offline stage, a library is built that includes five periodic gaits and a set of transition gaits that terminate in a periodic gait. The gaits are parameterized so as to allow a low-level joint controller to move the robot in a single step from a current pose to a desired final pose, with the desired pose planned in real-time at step initiation as a function of a terrain height map. In a similar vein, reference \cite{powell2012motion} first develops a set of feedback controllers for bipedal walking on flat ground, upstairs and downstairs, called motion primitives. In a second step, a set of feedback controllers is designed that evolve the robot from one motion primitive to another (termed motion transitions). The appropriate controller is selected at step transition. 

\noindent \textbf{Terrain Robust:} Previous locomotion work has also addressed gait controller design for a specified, finite set of terrain height perturbations \cite{ByTe09,dai2016planning,GriffinIJRR2016,xie2020learning}. During offline optimization, which could be via parameter optimization or reinforcement learning,  a ``score'' is assigned based on how the closed-loop system (consisting of the controller and robot) responds to a family of terrain profiles. The online controller is only allowed to use proprioception (such as IMU and joint encoder signals) to complete a locomotion task. In particular, the controller is not provided exteroceptive information on terrain profile, as in \cite{wu2010terrain} or \cite{powell2012motion}, for example. 

\noindent \textbf{MPC for Foot Placement without Terrain Preview:} In \cite{hamed2020quadrupedal}, the decoupled Linear Inverted Pendulum (LIP) model dynamics, first introduced in \cite{kajita20013d}, is used to solve a hybrid system-based optimization problem by computing center of pressure trajectories for a specified footfall pattern. These trajectories are computed at the beginning of each domain and are used as inputs to a virtual constraint-based Quadratic Program (QP) to realize joint torque commands. A separate hybrid system and MPC approach was performed in \cite{xiong2020global} for computing footfalls on a bipedal robot as well. The footfalls were chosen in order to minimize the error between the propagated LIP dynamics and a pre-specified reference trajectory.

\noindent \textbf{One Step-Ahead Prediction:} This paper adapts the philosophy of \cite{wurts2021collision} and designs an MPC controller that blurs the boundary between gait generation and trajectory planning. Our starting point is the one-step ahead gait controller in \cite{gong2021one,yukai2022arxiv}, which bridged the gap between the low-dimensional linear inverted pendulum (LIP) models in \cite{miura1984dynamic,kajita20013d,pratt2006capture,englsberger2011bipedal,WANGChevallereau2011,xiong2019orbit} and the method of virtual constraints and Hybrid Zero Dynamics in \cite{reher2016realizing, GriffinIJRR2016,da2019combining}. The main contribution of \cite{gong2021one,yukai2022arxiv} was to show that when the Center of Mass (CoM) dynamics of a physical robot are parameterized in terms of the \textit{angular momentum about the contact point instead of linear velocity}, the resulting model is only weakly affected by the angular momentum about the center of mass; in effect, the angular momentum about the contact point acts as a form of ``total momentum'', accounting for both linear momentum about the contact point and the angular momentum about the center of mass. When a 3D bipedal robot is controlled so that its CoM height is constant, a four-dimensional linear model (referred to as the 3D-ALIP), that is weakly perturbed by angular momentum about the center of mass, is extracted. When the perturbation term is dropped, the model simplifies to a pair of decoupled  2D models for the sagittal and frontal planes, respectively; see also the decoupled dynamics of the LIP for comparison \cite{kajita20013d}. 

The 2D ALIP models were subsequently used in \cite{gong2021one} to predict angular momenta at the end of the next step as a function of the robot's current angular momenta, position of its center of mass, and the swing-foot position at the end of the current step. When a (dead-beat) foot-placement controller was designed to place the swing foot so as to match the predicted angular momentum to desired angular momentum at the end of the next step, the Cassie bipedal robot was able to walk at 2.1 m/s, complete a 90$^o$ turn in 5 steps when walking at 1 m/s, and traverse significant slopes \cite{WaveFieldAutonomy2021}. 

Importantly, \textit{achieving this agile performance on Cassie required a skilled operator for the RC transmitter}, namely, an operator who has an intuitive feel for how rapid changes in commanded speed could result in workspace violations, self collisions (especially in lateral walking), foot slippages, and who could appropriately adjust foot clearance for locomotion over sloped terrain. This paper transforms the one-step ahead controller in \cite{gong2021one} into a multi-step horizon MPC controller. Moreover, the center of mass is allowed to move parallel to the ground, workspace constraints on the legs are included to avoid self collisions, and finally, the friction cone of the local terrain is incorporated. These contributions prepare the gait controller for integration with perception, mapping, and motion planning components illustrated in \cite{huang2021efficient}.   

\subsection{Contributions}

We provide the following contributions for enhancing terrain-aware locomotion:
\begin{itemize}
    \item Provide new insights about the exact CoM dynamics of a bipedal robot using CoM and angular momentum about the contact point as state variables. After applying a constraint to enforce the CoM height to remain a constant distance to the ground, we derive \textit{coupled} dynamics as opposed to the decoupled dynamics generated when using CoM velocity \cite{kajita20013d}. In \cite{kajita20013d} the decoupling is exact due to the point contact assumption, however additional assumptions are needed when using angular momentum. We justify why certain terms can be treated as negligible, which allows us to recover a decoupled linear system about the new state variables.
    \item Formulate an $N$-step receding horizon optimization problem that incorporates the 3D-ALIP dynamics and a piecewise linear terrain approximation for computing foot placements. The foot placement solutions, subject to workspace and approximate friction cone constraints, are computed in order to asymptotically achieve desired periodic trajectories at the end of the planning horizon.
    \item Create a novel set of virtual constraints and desired trajectories which can be used on a bipedal robot to achieve the desired motion of the 3D-ALIP dynamics for locomotion on piecewise linear terrain. The virtual constraints ensure that the CoM and swing toe remain parallel to the local ground plane, thereby extending the work in  \cite{gong2021one}, which assumed a constant ground height.
    \item Demonstrate the enhanced agility delivered by the control algorithm when implemented on a high degree-of-freedom (DoF) 3D Cassie bipedal robot, in comparison to previous results\cite{gong2021one}.
\end{itemize}

\section{3D Robot Models \& Dynamics}
\label{sec:3DModels}
\subsection{3D Physical Robot Model}
We assume a pinned point contact dynamic model of the form
\begin{equation}
    \label{eq:SwingPhasePinnedModel}
    D(q)\ddot{q} +  H(q,\dot{q}) = B(q) u,
\end{equation}
with no yaw motion about the stance foot (Cassie has a blade foot).
The generalized coordinates $q \in \real^{n_q}$, the vector of motor torques $u\in \real^{n_u}$, and the torque distribution matrix has full column rank. $D(q)$ is the mass inertia matrix, $H(q,\dot{q})$ is the combination of Coriolis, centrifugal, and gravity forces, and $B$ is the torque distribution matrix. For Cassie, $n_q=15$ due to the blade foot and $n_u=10$ if the ankle torque on the stance foot is included. In this work, it will be set to zero for simplicity, leaving $n_u=9$.

\subsection{Center of Mass Dynamics for Full-Order and Reduced-Order Model}
\begin{figure}
    \centering
    \includegraphics[width=0.27\textwidth]{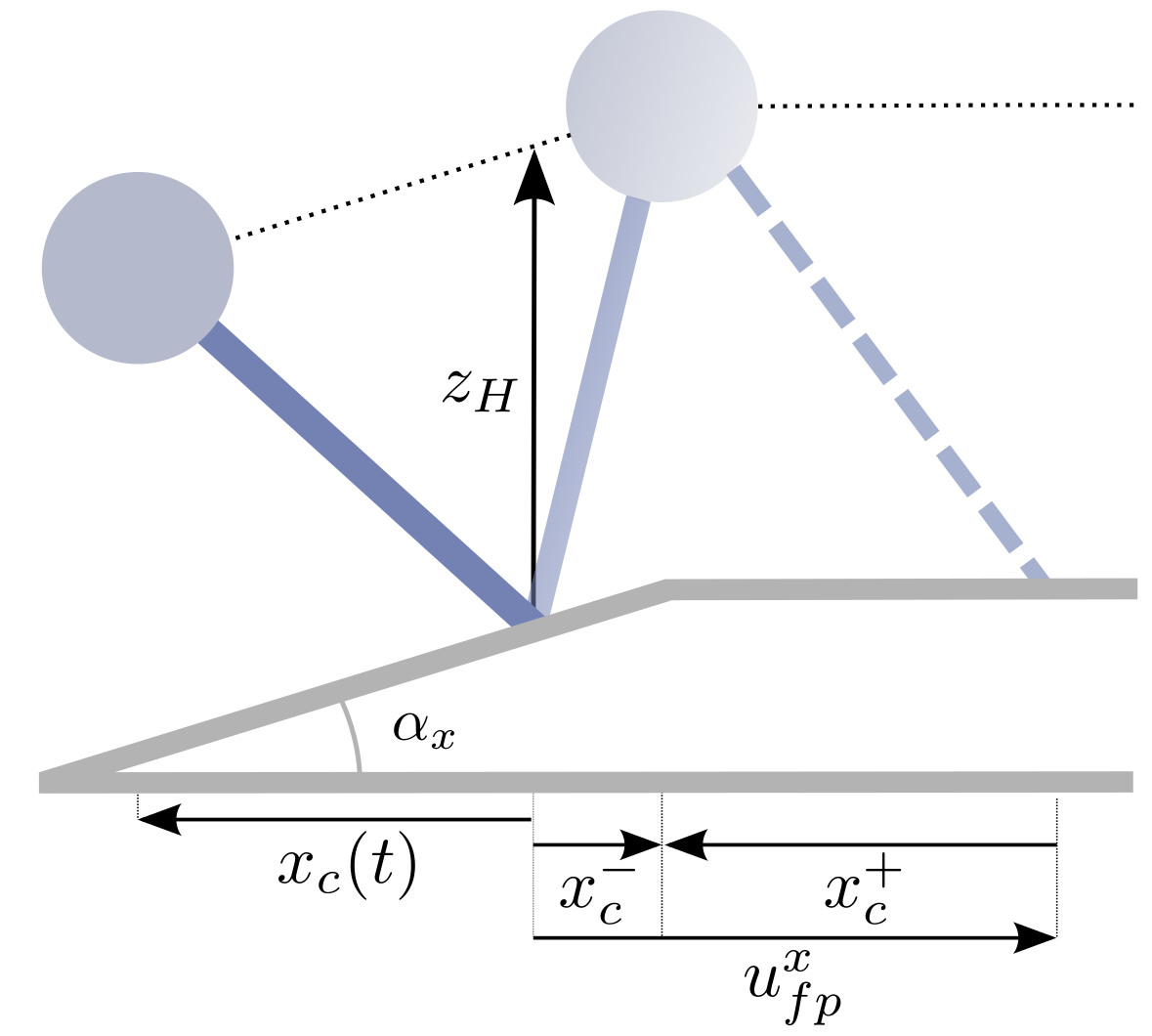}
    \caption{The planar linear inverted pendulum is shown. $x_c(t)$ represents the location of the center of mass with respect to the stance foot at time $t$. The control input $u_{fp}^x$ denotes the foot placement at the end of the current step. The state before and after the instantaneous impact is denoted with a minus (-) and a plus (+) sign, respectively.}  
    \label{fig:2d_lip}
\end{figure}

For a 3D robot with a point contact, the dynamics for CoM positions and angular momenta about the contact point can be written as follows,
\begin{equation}
    \label{eqn:ZDnaturalCoordinates}
    \begin{aligned}
      \dot{x}_c&= \frac{L^y}{m z_c} + \frac{\dot{z}_c}{z_c}x_c - \frac{L^y_c}{m z_c}\\
      \dot{y}_c&= -\frac{L^x}{m z_c} + \frac{\dot{z}_c}{z_c}y_c + \frac{L^x_c}{m z_c}\\
      \dot{L}^x&= - m g y_c \\
      \dot{L}^y&=m g x_c.
    \end{aligned}
\end{equation}
where $m$ is the mass of the robot, $g$ is the gravitational constant, $x_c,y_c,z_c$ denotes the CoM position with respect to the contact point, $L^{x,y, z}$ denotes the angular momenta about the $x,y,z$-axes of the contact point, and $L_c^{x,y, z}$ denotes the angular momentum about the center of mass. Assuming $p = [x_c,\ y_c,\ z_c]^T$ and $F_g$ is the gravitational force vector applied to the center of mass, the above can be derived from the fact that $L = p \times m \dot p + L_c$ and $\dot L = p \times F_g$. 

Motivated by the CoM height constraint from the LIP model \cite{kajita2003biped}, we will later design virtual constraints to impose the following relations on the evolution of the center of mass,
\begin{equation} \label{eq:zc_constr}
    \begin{split}
        z_c &= k_x x_c + k_y y_c + z_H \\
        \dot z_c &= k_x \dot x_c + k_y \dot y_c,
    \end{split}
\end{equation}
under which the model becomes 
\begin{equation}
    \label{eqn:ZDnaturalCoordinatesPre}
    \begin{aligned}
      \dot{x}_c&= \frac{L^y}{m z_H} + \frac{k_y}{z_H}(x_c\dot{y}_c - y_c\dot{x}_c) - \frac{L^y_c}{m z_H}\\
      \dot{y}_c&= -\frac{L^x}{m z_H} - \frac{k_x}{z_H}(x_c\dot{y}_c - y_c\dot{x}_c) + \frac{L^x_c}{m z_H}\\
      \dot{L}^x&= - m g y_c \\
      \dot{L}^y&=m g x_c.
    \end{aligned}
\end{equation}
The slope of the $x-y$ ground plane is represented by $k_{x} = \tan \alpha_{x}$ and $k_{y} = \tan \alpha_{y}$, respectively. The constant ground height parameter is represented by $z_H$  (as shown in Fig. \ref{fig:2d_lip}). By the equation $L = L_c  + [x_c,y_c,z_c]^T \times m ~ [\dot x_c, \dot y_c, \dot z_c]^T$, the above can be rewritten to make $L^z$ and $L_c$ explicit,
\begin{equation}
    \label{eqn:ZDnaturalCoordinatesPost}
    \begin{aligned}
      \dot{x}_c&= \frac{L^y}{m z_H} + \frac{k_y}{m z_H}(L^z - L^z_c) - \frac{L^y_c}{m z_H}\\
      \dot{y}_c&= -\frac{L^x}{m z_H} - \frac{k_x}{m z_H}(L^z - L^z_c) + \frac{L^x_c}{m z_H}\\
      \dot{L}^x&= - m g y_c \\
      \dot{L}^y&=m g x_c.
    \end{aligned}
\end{equation}

In \cite{gong2021one,yukai2022arxiv}, it has been shown that both $L^{x}_c$ and $L^{y}_c$ are small compared to $L^{x}$ and $L^{y}$, respectively. By using $L^{x}$ and $L^{y}$ as state variables in place of the CoM velocities, neglecting $L_c$ has only a small effect on the dynamic accuracy during normal walking, even for robots with \textit{heavy} legs. 

Next, we make the case that $(L^z - L_c^z)$, which is the same as $(x_c\dot{y}_c - y_c\dot{x}_c)$, can be neglected. Some readers might already believe $(L^z - L_c^z)$ is a small term. For others, there are two ways to look at it intuitively: 1) When a robot is walking purely longitudinally ($y_c = \dot{y}_c = 0$) or laterally ($x_c = \dot{x}_c = 0$) the product is zero. For diagonal movement, we can define a new frame aligned with the walking direction, which also makes the $y_c$ and $\dot y_c$ terms small. 2) If we project the position vector and velocity vector to a horizontal plane, then $(x_c\dot{y}_c - y_c\dot{x}_c)$ is the cross product of these two projected vectors. Throughout a walking gait, either the angle between the two projected vectors is small or the magnitude of the projected position vector is small. \textit{Note:} Similar approximations are not needed in \cite{kajita20013d} because the coupling term ($x_c\ddot{y}_c - y_c\ddot{x}_c$) can be nullified by the assumption of a point contact (zero torque applied). 

As a result of these approximations, we arrive at the dynamics for the 3D-ALIP model with center of mass evolving as in \eqref{eq:zc_constr},
\begin{equation}
    \label{eq:ALIP_angular}
    \dot{\mathbf{x}} = \begin{bmatrix}
    \dot{x}_c \\ \dot{y}_c \\
    \dot{L}^x \\ \dot{L}^y
    \end{bmatrix} = \underbrace{\begin{bmatrix}
    0 & 0 & 0 & \frac{1}{m z_H} \\
    0 & 0 & -\frac{1}{m z_H} & 0 \\
    0 & - m g & 0 & 0 \\
    m g & 0 & 0 & 0
    \end{bmatrix}}_{A} \underbrace{\begin{bmatrix}
    x_c \\ y_c \\ L^x \\ L^y
    \end{bmatrix}}_{\mathbf{x}}.
\end{equation}
 While the model \eqref{eq:ALIP_angular} is ultimately the same as in \cite{gong2021one,yukai2022arxiv}, these references do not provide a derivation of the model \eqref{eqn:ZDnaturalCoordinatesPre} and \eqref{eqn:ZDnaturalCoordinatesPost} nor do they explain the simplifications needed to arrive at \eqref{eq:ALIP_angular}.
We reiterate that the state includes the angular momenta rather than the center of mass velocities; the benefits of this selection have been highlighted in several related publications \cite{sano1990realization,GriffinIJRR2016,powell2016mechanics_2,gong2021one}. 

\subsection{Foot Placement as a Control Variable}
The dynamic model \eqref{eq:ALIP_angular}, which describes the evolution of the centroidal dynamics when the robot is in single support, is not affected by the motor torques\footnote{Recall that we are leaving the stance ankle passive in this study.}. So how to control it? As in  \cite{RA86a,HORA91,powell2016mechanics_2,gong2021one}, we use the placement of the end of the swing leg as a step-to-step actuator. Under conservation of angular momentum and \eqref{eq:zc_constr}, if $\mathbf{x}^-$ is the solution of \eqref{eq:ALIP_angular} just before impact, and $\mathbf{x}^+$ is the value of the state just after the (instantaneous) impact, then 
\begin{equation} 
    \label{eq:lip_with_ufp}
    \mathbf{x}^+ =\mathbf{x}^- + B \mathbf{u}_{fp},
\end{equation}
where 
\begin{equation}
    B = \begin{bmatrix} -1 & 0 & 0 & 0 \\ 0 & -1 & 0 & 0 \end{bmatrix}^T
\end{equation}
and $\mathbf{u}_{fp}$ is the foot placement (or resultant vector emanating from the current stance contact point to the desired swing foot impact location) and $B$ represents the instantaneous change in coordinate systems to the new stance foot.

We assume that the height of the swing leg is controlled so that the duration of each step of the robot is fixed at $T_s$. Hence, during the $k$-th step of the robot, the height of the swing leg above the ground is regulated to be positive for $ (k-1) T_s < t < k T_s$ and zero at $k T_s$. The relative $(x, y)$ position of the swing leg end at time $k T_s$ is selected so as to achieve a desired evolution of \eqref{eq:ALIP_angular} for $t > k T_s$. Fig.~\ref{fig:2d_lip} shows a planar schematic of the $x$-component of the 3D-ALIP.

\noindent \textbf{\textit{Remark:}} Reduced-order models similar to \eqref{eq:ALIP_angular} and \eqref{eq:lip_with_ufp} were used in \cite{xiong2020global} for a foot placement controller based on the LIP model, which is parameterized by CoM velocity instead of angular momentum.  Moreover, the models for the impact dynamics are based on the same conservation of angular momentum assumption. Nevertheless, our use of angular momentum is not a ``matter of taste''; it is the better state choice for controlling a variety of bipedal robots as explained in \cite{yukai2022arxiv}: (a) the CoM velocity is more sensitive to motor torque transients because it is relative degree one as opposed to angular momentum, which is relative degree three; (b) linear velocity does not capture the natural interchange between linear momentum and angular momentum about the center of mass; and (c), there are typically large jumps in the velocity variables at impact for fast moving bipedal robots. To predict this jump, all states of the robot and the impact model need to be known, which makes the online prediction infeasible in practice. Therefore, an assumption of continuity is widely adopted. This is much less of an issue for angular momentum about the contact point because it is invariant to the impulse generated at contact. Hence, even the impact model functions better in the angular momentum coordinates.

\section{MPC Formulation for Foot Placement Control of 3D-ALIP} \label{Sec:MPC_Formulation}
Our next goal is to compute desired footholds for the robot to execute in order to converge to some desired (centroidal) state at the end of each step. In \cite{gong2021one}, a single $\mathbf{u}_{fp}$ was selected step-to-step to achieve a desired value of the angular momenta in \eqref{eq:ALIP_angular} at the end of the next step. The values of $x_c$ and $y_c$ were not regulated and constraints on self-collisions and friction cone were ignored. The one-step ahead foot placement method in \cite{gong2021one} can be viewed as MPC with a terminal linear equality constraint on a portion of the state vector and no inequality constraints. Here, we will develop an MPC formulation of foot placement control over a multi-step horizon, a quadratic cost to be minimized, and appropriate linear inequality constraints to avoid self collisions in order to render the solution terrain adaptive.

We begin by defining
\begin{itemize}
    \item $N_s$, the number of (robot) steps of (fixed) duration $T_s$ in the MPC control horizon.
    \item $\delta t=T_s / N_{\delta t}$ is the sample period for the controller, where $N_{\delta t}>1$ is an integer.
    \item $A_{\delta t} := \exp(A \delta t)$, the state transition matrix of \eqref{eq:ALIP_angular} for a time duration of $\delta t$ seconds. 
    \item $\mathbf{x}_0$ in the MPC will always be the predicted solution  $\widehat{\mathbf{x}}(kT_s,t)$ of \eqref{eq:ALIP_angular} just before impact of the current step, based on the measured value of state at time $t$, that is,
\begin{equation}
\label{eq:DefX0}
\mathbf{x}_0:= \widehat{\mathbf{x}}(kT_s,t):= \exp(A(kT_s-t)) \mathbf{x}(t).
\end{equation}
    This will allow us to work with a fixed control horizon.
    \item Next, we define the discrete-time dynamics 
\begin{equation}
\label{eq:InstraStepDynamics}
  \mathbf{x}_{i+1} = \begin{cases}
        A_{\delta t} (\mathbf{x}_i + B \mathbf{u}_{fp,i}), & i = j N_{\delta t}, \\
        & \hfill  0 \le j \le (N_s-1)  \\
        A_{\delta t} \mathbf{x}_i, & \texttt{otherwise}
    \end{cases}
\end{equation}
for use in our MPC problem as it will allow us to place constraints on the intra-step evolution of $x(t)$, that is, its behavior between steps.
\item $\mathbf{x}_i^{\rm des}$ is the desired evolution of the state and the associated error term is
\begin{equation}
    \label{eq:StateErrorMPC}
    \mathbf{x}_{e,i} := \mathbf{x}_i - \mathbf{x}^{\rm des}_i. \\
\end{equation}
\end{itemize}

\subsection{MPC Problem}
An $N_s$-step horizon MPC control problem with quadratic cost and linear constraints can now be formulated as 
\begin{equation} \label{eq:mpc}
\begin{split}
    &\min_{\mathbf{U}_{fp}} J = \sum_{i= 0}^{N_{\delta t} N_s-1} \mathbf{x}_{e,i}^T Q_i \mathbf{x}_{e,i} + \mathbf{x}_{e,N_{\delta t} N_s}^T Q_{f} \mathbf{x}_{e,N_{\delta t} N_s} \\
    &\mathbf{subject\ to} \\
    & \eqref{eq:DefX0}, \eqref{eq:InstraStepDynamics}, \text{ and } \eqref{eq:StateErrorMPC}\\
    &\forall\ \mathbf{x}_i \in \mathcal{X} \text{ and } \forall\ \mathbf{u}_{fp,i} \in \mathcal{U},
\end{split}
\end{equation}
where $\mathbf{U}_{fp} = \left[\mathbf{u}_{fp,0}, \mathbf{u}_{fp,N_{\delta t}}, \ldots, \mathbf{u}_{fp,N_{\delta t}(N_s-1)}  \right]$. The solution returns the optimal foot placement sequence. Only the first value $\mathbf{u}_{fp,0}$ in the sequence is applied. The state and control constraint sets, $\mathcal{X}$ and $\mathcal{U}$, respectively, are defined in the following subsection. 

\subsection{Constraint Sets}
The state constraint set $\mathcal{X}$ is the union of the mechanical safety set $\mathcal{X}^{\rm mech}$ and the friction cone safety set $\mathcal{X}^{\rm slip}$. The foot placement safety set $\mathcal{U}$ is also used to prevent foot collisions, most importantly in the lateral direction. The mechanical safety and foot placement safety sets are constructed as box constraints related to the geometrical limitations of the robot. The ground friction cone (based on the 3D-ALIP dynamics) is used to constrain the intra-step CoM positions. 

We begin by deriving the constraint for $x_c$ assuming ground slope purely in the sagittal direction. In this model, the ground reaction force is applied collinearly through the contact leg because the point mass moves parallel to the ground plane. Taking advantage of the known slope of the terrain, we derive the resultant tangent and normal forces with respect to the ground to be \begin{equation} \label{eq:GRF_rotate}
    \begin{bmatrix}
        F_{T_x} \\ F_{N_x}
        \end{bmatrix} =\begin{bmatrix}
        \cos \alpha_x & \sin \alpha_x \\ 
        -\sin \alpha_x & \cos \alpha_x \\
        \end{bmatrix} \begin{bmatrix}
        F_x \\ F_z
        \end{bmatrix}.
\end{equation}

Given the defined motion constraints, we compute the relative force ratios
\begin{equation} \label{GRF ratios}
    \begin{split}
        F_{x/z} &= \dfrac{F_x}{F_z} = \dfrac{x_c}{k_x x_c + z_H} 
    \end{split}
\end{equation}
and combine this with a Coulomb static friction constraint ($|F_{T_{x}}| \leq \mu F_{N_{x}}$) to compute the slip constraint on $x_c$,
\begin{equation} \label{eq:slip}
    |F_{x/z} + k_x| \ \leq  -\mu k_x F_{x/z} + \mu.
\end{equation}
For known $k_x$ and $\mu$ (friction coefficient), the explicit constraint is given as
\begin{equation} \label{eq:slip_limit_simple}
    |x_c| \ \leq \frac{(\mu - k_x)z_H}{1+k_x^2} = x_c^{\rm slip}.
\end{equation}
For ground slope purely in the lateral direction and known $k_y$, a similar constraint on $y_c$ can be derived.

We emphasize that this is not an exact friction slip constraint due to the simplifications of using the 3D-ALIP for ground reaction force estimation. It can, however, be combined with an under-approximation of the friction coefficient $\mu$ to enable safer foot placements. 
For example, multiplying $\mu$ by $\frac{1}{\sqrt{2}}$ in \eqref{eq:slip_limit_simple} results in a linearized under-approximation of the Coulumb cone.

\subsection{Cost Design}
The cost function is the sum of a running cost with non-zero weights at step transitions, that is, $Q_i = \mathbf{0}, \forall i \notin \{N_{\delta t},  \ldots, (N_s-1) N_{\delta t} \}$ and a terminal cost $Q_f$. Given a desired longitudinal angular momentum $L^{y,\rm des}$ and step width $W$, we can use the solutions of \eqref{eq:ALIP_angular} to compute the desired state of the corresponding 2-step periodic orbit for the corresponding stance foot (by following \cite{gong2021one}).
With the assumption of conservation of angular momentum about the contact point, we substitute $L^{y, \rm des} = L^y(0) = L^y(T_s)$, $L^x(0) = -L^x(T_s)$, and $y_c(0) = \nicefrac{W}{2}$ into the trajectory solutions of \eqref{eq:ALIP_angular} and solve the resultant linear system of equations. The resultant desired state at each impact is
\begin{equation}
    \label{eq:des_states}
    \mathbf{x}^{\rm des}_i = \begin{bmatrix}
        \frac{1}{m z_H \ell} \tanh(\nicefrac{\ell T_s}{2}) L^{y,\rm des} \\
        -\frac{1}{2}\sigma W \\
        \frac{1}{2} \sigma  m H \ell W \tanh(\nicefrac{\ell T_s}{2}) + L^{x,\rm offset} \\
        L^{y, \rm des}
    \end{bmatrix},
\end{equation}
where $\ell = \sqrt{\nicefrac{g}{z_H}}$, $L^{x,\rm offset}$ is an additional lateral angular momentum term, and $\sigma$ is +1 for left stance and -1 for right stance. Without $L^{x,\rm offset}$ the controller will walk nominally with zero lateral velocity. 

The terminal cost $Q_f$ is computed as the optimal cost-to-go of a Discrete-Time Algebraic Ricatti Equation for a periodic 2-step trajectory including impact, combining \eqref{eq:ALIP_angular} and \eqref{eq:lip_with_ufp}, and ignoring constraints. This selection of terminal cost ensures recursive feasibility via Bellman's principle of optimality \cite{borrelli2017_LMPC}.

\section{Virtual Constraints and Foot Placement Implementation on Cassie} 
\label{sec:virtualconstraints}
The computed foot placement solution is implemented on the high DoF bipedal robot Cassie Blue through the use of specially designed virtual constraints. As documented in \cite{gong2021one}, an important feature of the 3D-ALIP model is that the mass of the swing leg and its corresponding momentum are accounted for in $L^x$ and $L^y$. 

Cassie is a 32 kg, 20 DoF biped robot actuated at ten joints. Each leg has seven joints, five of which are actuated while the remaining two are constrained by springs \cite{Yukai2018}. To achieve a desired foot placement, we must define the control variables and generate their reference trajectories. The nine control variables $h$ and the corresponding references $h_d$ are defined as follows:
\begin{equation}
    h = 
    \begin{bmatrix}
    \rm{torso\ pitch} \\
    \rm{torso\ roll} \\
    \rm{stance\ hip\ yaw} \\
    \rm{swing\ hip\ yaw}\\
    p_{{\rm CoM_{\rm proj}} \to {\rm CoM}}^z \\
    p_{\rm st \to \rm sw}^x \\
    p_{\rm st \to \rm sw}^y \\
    p_{\rm st \to \rm sw}^z \\
    \rm{absolute\ swing\ toe\ pitch}
    \end{bmatrix}
\end{equation}
The desired reference trajectories are parametrized by a time-based phase variable $s = \nicefrac{(T_s-t)}{T_s}$ where $t$ is the time since last impact. We set the reference values for torso pitch and torso roll to be zero. To enable turning, one-half of the total desired turn angle $\Delta \psi$ at step end is set as the reference position for both stance and swing yaw motor joints at the end of the current step \cite{gong2021one}. The reference absolute swing toe pitch angle is adjusted to align with the given terrain slope. $p_{\rm{st} \to \rm{sw}}$ is the position vector of the swing leg toe relative to the stance leg toe and $p_{\rm{CoM}_{\rm proj} \to \rm{CoM}}^z$ represents the constant height parameter $z_H$, between the center of mass and an \textit{inclined} ground. 
\begin{equation}
    h_d(s) :=
    \begin{bmatrix}
    0 \\
    0 \\
    (1-s) h_3^{\rm init} - s(\frac{1}{2} \Delta \psi) \\
    (1-s) h_4^{\rm init} + s(\frac{1}{2} \Delta \psi) \\
    z_H \\
    \frac{1}{2}[(1+\cos(\pi s)) h_6^{\rm init} + (1-\cos(\pi s)) p_{\rm st \to \rm sw}^{x,\rm des}] \\
    \frac{1}{2}[(1+\cos(\pi s))h_7^{\rm init} + (1 - \cos(\pi s)) p_{\rm st \to \rm sw}^{y,\rm des}] \\
    \beta_1 s^2 + \beta_2 s + \beta_3\\
    k_x
    \end{bmatrix}.
\end{equation}
\noindent \textbf{\textit{Remark:}} The virtual constraints in \cite{gong2021one} regulate the CoM to remain at a constant height with respect to the pinned stance foot and do not account for the CoM height constraint used in this paper.  We instead derive a new kinematic relation $p_{\rm CoM_{\rm proj} \to \rm{CoM}}^z$ that computes the height of the CoM relative to the projected position on the terrain. More explicitly, $p_{\rm CoM_{\rm proj} \to \rm{CoM}}^z = p_{\rm st \to \rm CoM}^z - k_x p_{\rm st \to \rm CoM}^x - k_y p_{\rm st \to \rm CoM}^y$. The desired swing toe angle is modified to align with the ground.

Outputs, $p_{\rm st \to \rm sw}^{x,y}$ are set equal to the MPC foot placement solution $\mathbf{u}_{fp,0}$ described in \eqref{eq:mpc}. The z component of $p_{\rm st \to \rm sw}$ can be easily computed with the knowledge of $k_x$ and $k_y$.  We use sinusoidal references for the $x$ and $y$ components (following \cite{gong2021one}) and a parabola for the $z$ component parametrized by the initial and final heights of the swing leg determined by foot placement and the relative time and height of a user-defined step clearance (The $\beta_i$ parameter derivations are omitted due to space constraints). $h_i^{\rm init}$ denotes the value of each output at the beginning of each new step. 

We implement an inverse kinematics and passivity-based control schema to track these constraints on the physical robot (see \cite{gong2021one,yukai2022arxiv}).

\section{Results} \label{Sec:ExperimentResults}
This section illustrates the capabilities of the new ALIP-MPC controller in simulation and experiments.\footnote{Open-source code can be found at \href{https://github.com/UMich-BipedLab/cassie_alip_mpc}{https://github.com/UMich-BipedLab/cassie\_alip\_mpc}.} It also compares the new controller with a previous one-step-ahead (or ALIP-1step) controller implemented on Cassie in \cite{gong2021one}.

\subsection{Simulation Results}
Simulation is used to demonstrated the advantages of the proposed gait ALIP-MPC controller with respect to the ALIP-1step controller in \cite{gong2021one}. The 20 DoF simulation model and the ALIP model are identical in both cases, except that the ALIP model used in the MPC controller takes the slope into account, following general motion constraints in \cite{kajita20013d}. 

\begin{figure}
    \centering
    \includegraphics[width=\columnwidth]{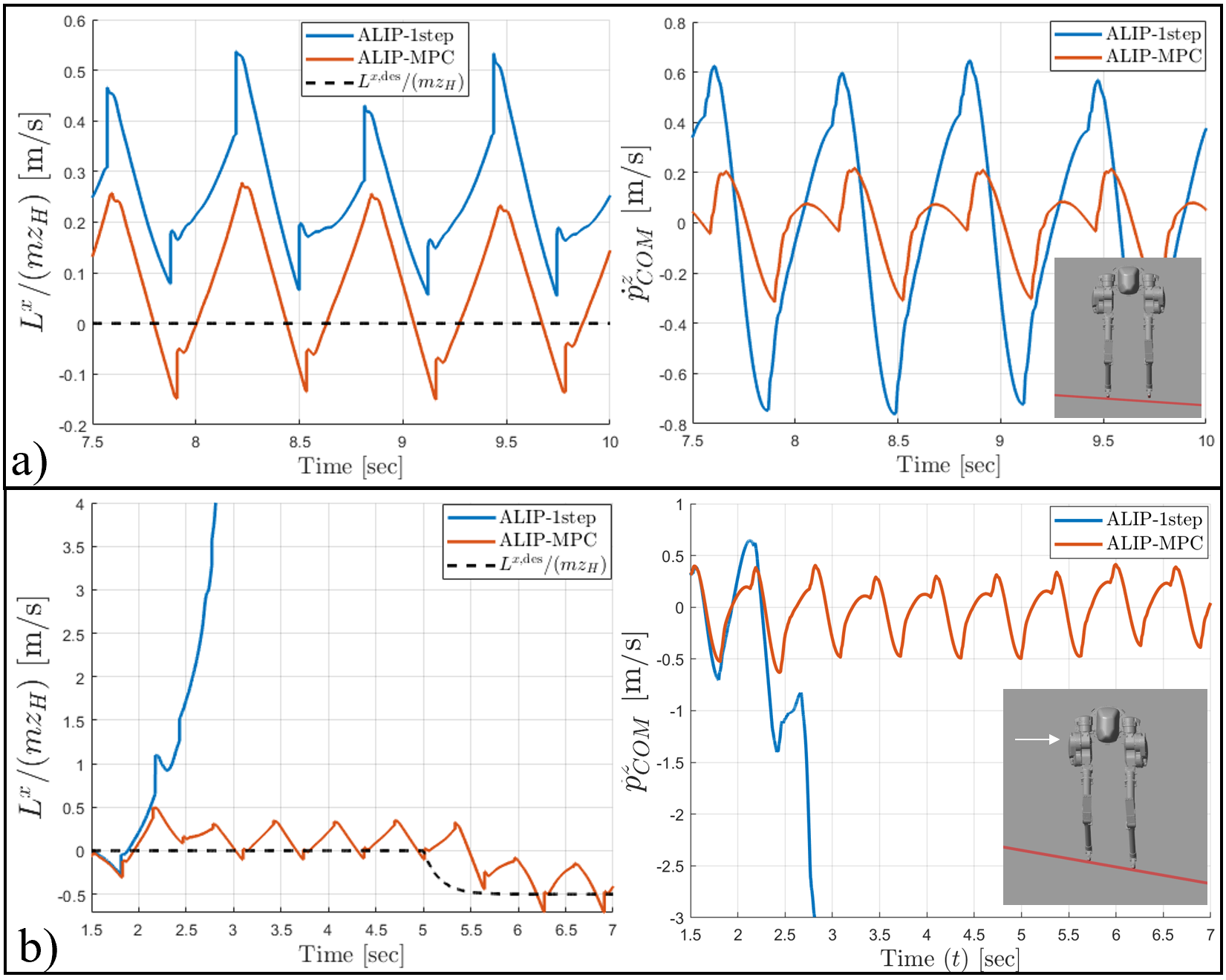}
    \caption{Comparison between proposed ALIP-MPC controller and ALIP-1step (one-step-ahead) controller \cite{gong2021one} on inclines. (a) The ground has a  $5^o$ lateral slope and Cassie is commanded to walk with zero velocity. The lack of slope information in the ALIP-1step controller leads to an increase in magnitude of the CoM velocity and increased tracking error of the constant CoM height assumption. In (b) a ground incline of $11^o$ causes the ALIP-1step controller to fail, while the ALIP-MPC controller allows Cassie to walk downhill laterally at approximately 0.5 m/s.}
    \label{fig:lat_slope_compare}
\end{figure}

\begin{figure}
    \centering
    \includegraphics[width=\columnwidth]{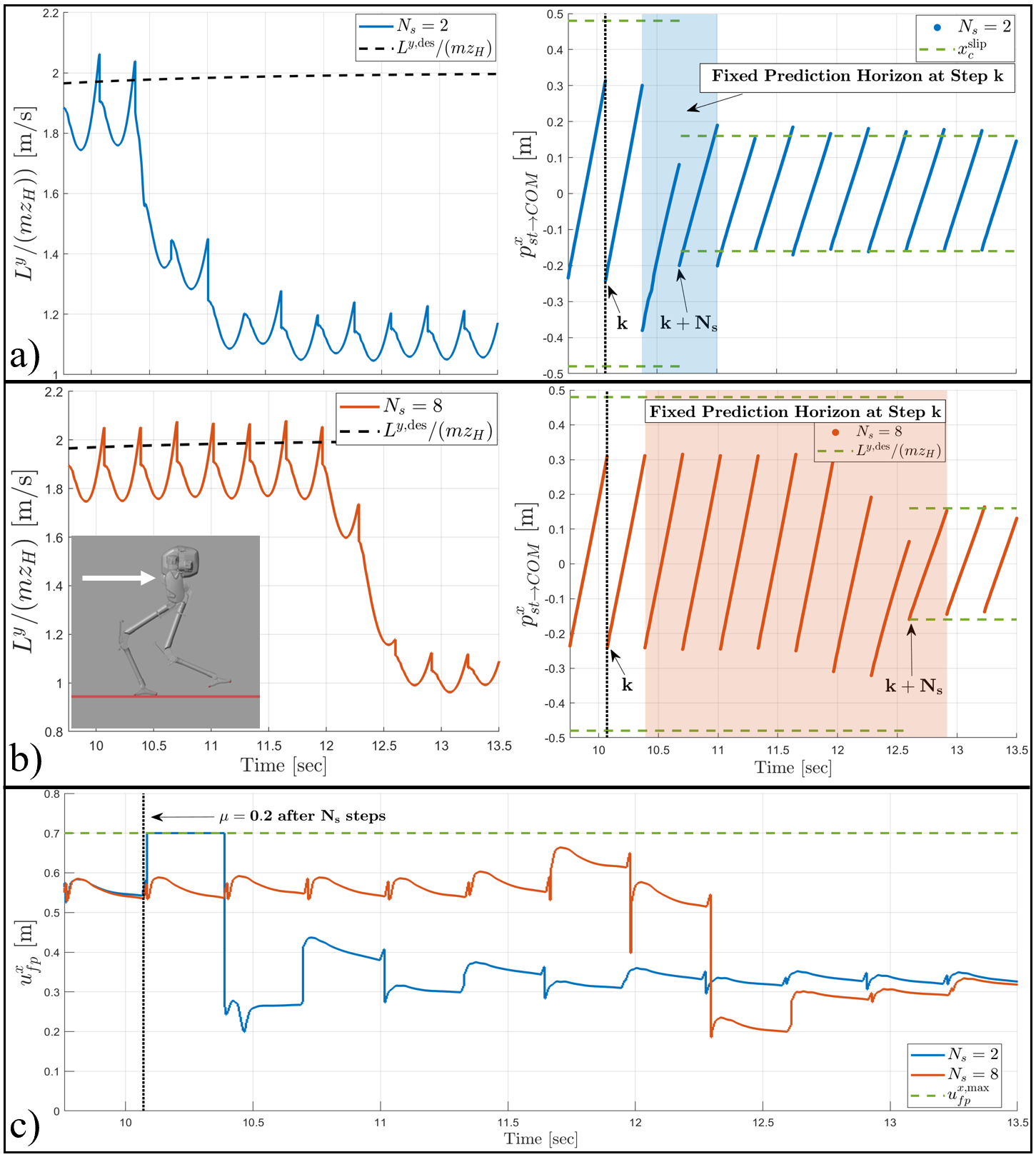}
    \caption{Comparison between (a) the 2-step horizon and (b) the 8-step horizon implementation of the proposed ALIP-MPC controller. At $t = 10.1$, both controllers are informed that the friction coefficient will reduce to 0.2 at the end of their respective horizons. In c) we compare the MPC foot placement solutions calculated during each step for both planning horizon choices. The 2-step version is restricted from extending its swing leg too far which results in a friction violation at step $k+N_s$. The 8-step version has more time to reduce the robot's velocity in order to satisfy the constraint, and only uses one additional step to affect the change. The slight friction cone violations are from the ALIP approximation, which re-enforces the need of an under-approximated friction coefficient.}
    \label{fig:frictionSim}
\end{figure}

Fig.~\ref{fig:lat_slope_compare} highlights the importance of the ALIP model and modified virtual constraints accounting for slopes in the lateral plane. On terrain with a $5^o$ lateral slope (Fig.~\ref{fig:lat_slope_compare}a), the MPC controller achieves an average lateral velocity close to the desired zero velocity reference because it adapts to the slope, while with the one-step-ahead controller, the robot drifts downhill. The drift is caused by untimely impacts and unbalanced forces on the uphill vs downhill contacts. The virtual constraints of each controller are designed to zero the vertical velocity of the CoM. With the MPC controller, $v_z$ has only a small oscillation caused by imperfect low-level tracking on the 20 DoF model. Walking laterally at 0.5 m/s is compared in Fig.~\ref{fig:lat_slope_compare}b, with similar results. In short, the improvements over \cite{gong2021one} allow walking over steeper terrain. 
Fig.~\ref{fig:frictionSim} illustrates how changing the length of the prediction horizon affects the ability of the ALIP-MPC controller to satisfy constraints. The friction parameter is modified online and we confirm the benefits of using a larger horizon for safer walking when imposing step length restrictions $u_{fp}^{\rm max}$.
\noindent \textbf{\textit{Remark:}} The benefits of including slope incline within the proposed ALIP-MPC controller is twofold. Firstly, the constant height constraint is embedded within the 3D-ALIP model, instead of the flat ground assumption, to more accurately approximate the tangential contact forces. We then formulate an MPC problem with constraints imposed on these forces to compute foot placements, which prevent slipping. Secondly, we modify the virtual constraints of \cite{gong2021one} such that they include slope data. Fig. \ref{fig:lat_slope_compare} highlights the benefits of this improvement to reduce CoM height variation and untimely contacts, leading to improved velocity tracking.

\subsection{Experimental Results}
The proposed ALIP-MPC controller was coded in C++ and run on a secondary computer in a Linux environment. The planning horizon of the controller was set to four steps ($N_s = 4$) with a step period of 0.3 seconds and an intra-step time discretization of 10 ms. Foot placement updates were sent over UDP to the primary computer on Cassie at 250 Hz. The resultant QP was code-generated using CasADi \cite{Andersson2019} and evaluated using a primal-dual active set algorithm \cite{ANDERSSON2018_qrqp}. 

\begin{figure}
\centering
\includegraphics[width=0.65\columnwidth]{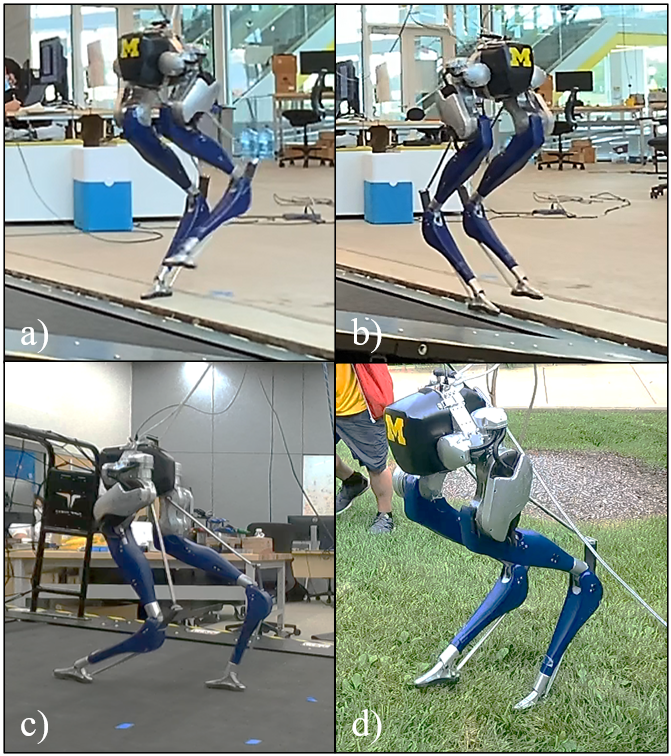}
\caption{Images from a variety of experiments performed with Cassie Blue using the gait controller discussed in this paper. (a) Forward walking at 1 m/s with a transition onto a stationary treadmill inclined at 13$^o$. (b) Lateral walking at 0.5 m/s with a transition onto a stationary treadmill inclined at 13$^o$. (c) Forward walking at 1.5 m/s on a moving treadmill inclined at 6$^o$. (d) Forward walking at 1 m/s on a wet, grassy slope inclined at 22$^o$. Videos available at \cite{gibson2021_youtube}.}
\label{fig:ExperimentResult}
\end{figure}
\begin{figure*}
    \centering
    \includegraphics[width=0.82\textwidth]{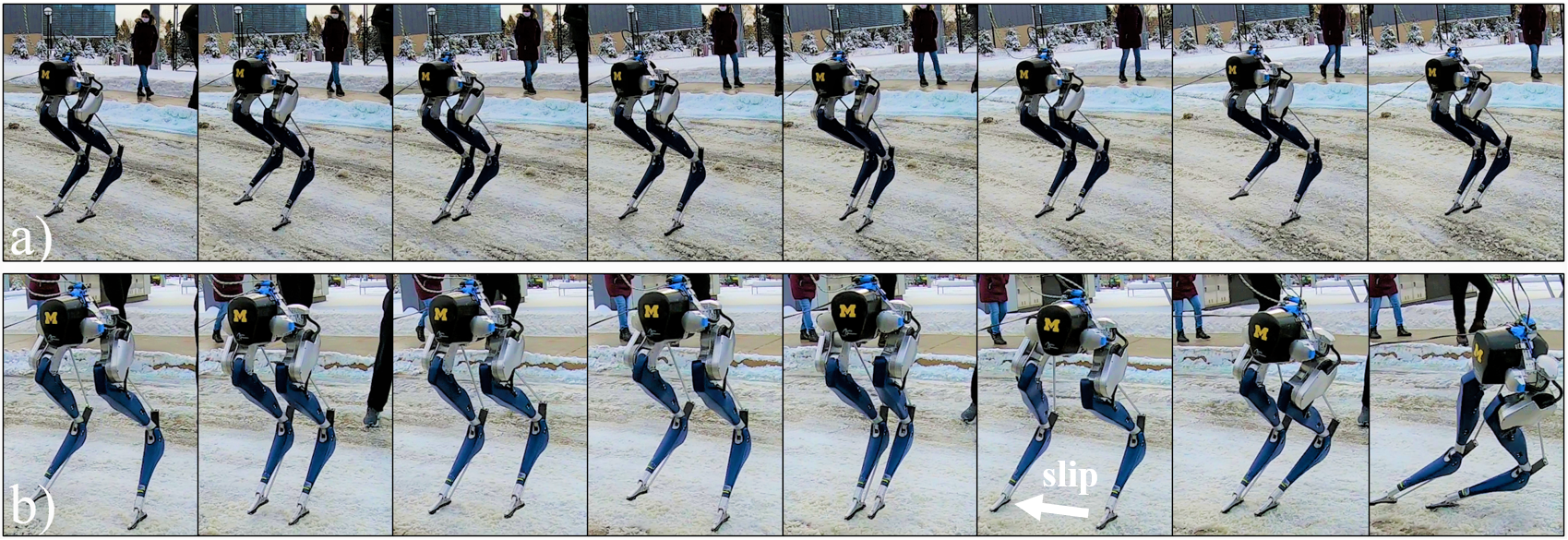}
    \caption{Lateral walking (desired 1 m/s) on snow/ice mixture for friction coefficients (a) $\mu = 0.6$ and (b) $\mu = 1.0$. The slip constraint is embedded in the intra-step dynamics of the MPC formulation. In (b) Cassie slips when the magnitude of the lateral CoM component is too large whereas in (a), the more cautious friction coefficient properly restricts the foot position, and thereby limits the robot's speed.}
    \label{fig:snow_lat_comp}
\end{figure*}

The MPC controller was implemented on Cassie Blue and evaluated in a variety of situations shown in Fig.~\ref{fig:ExperimentResult} and \ref{fig:snow_lat_comp}.
\vspace*{-.1cm}
\begin{itemize}
    \item \textbf{Inclined walking.} Cassie walked forward on a treadmill inclined at 6$^o$ at a maximum speed of $1.5$ m/s, and also walked laterally at a maximum speed of 0.5 m/s on a stationary treadmill inclined at 13$^o$. When we tried the lateral walking with the one-step ahead controller of \cite{gong2021one}, which does not actively constrain the workspace of the legs, the robot tripped and fell. We test the limits of our control method by having Cassie Blue walk up an uneven slope with an estimated average incline of 22$^o$. Cassie was successful in walking up the slope in the sagittal direction at 1 m/s and the lateral direction at 0.3 m/s. Speed in the lateral direction is inherently slower due to hardware limits on step width.
    
    \item \textbf{Transitioning from Flat Ground to an Incline.} In this experiment, an operator sent the slope information to Cassie at the transition. Performance relied heavily on the timing and accuracy to which the operator was able to switch the estimated ground slope with respect to the body frame of the robot. While the transitions were not ideal, due to operator error, a noticeable improvement of maintaining a constant CoM height with respect to the ground was seen compared to \cite{gong2021one}. Future work will remove operator dependence and integrate with perception and \cite{huang2021efficient}.
    
    \item \textbf{Rapid Changes in Lateral Velocity.}  In order to validate the ability of the MPC controller to achieve self-collision constraints, as in  Sect.~\ref{Sec:MPC_Formulation}, we constrain the lateral foot placement solution to remain within the safety set $\mathcal{U}$ for all experiments. As shown in \cite{gibson2021_youtube}, the swing legs avoid collisions when rapidly changing the lateral target velocities.

    \item \textbf{Avoid Slipping on a Snow-Ice Mixture.} Via the RC transmitter, the operator adjusted the assumed friction coefficient to prevent slipping on the snow-ice mixture shown in Fig.~\ref{fig:snow_lat_comp}. With poor estimation (or omission) of the $x_c^{\rm slip}$ constraint, the robot slips and falls.
\end{itemize} 

\section{Conclusions and Future Work} \label{Sec:Conclusions} This paper extended a controller based on a constant CoM height and a one-step ahead prediction of angular momentum \cite{gong2021one} in three ways. First, the 3D-ALIP model was derived to allow the robot's center of mass to exhibit piecewise planar motion. The resulting model, while similar to the LIP in~\cite{kajita20013d}, more closely models a physical robot. Second, a four-step ahead MPC controller was provided, which, importantly, allowed realistic workspace and terrain-centric constraints in the MPC formulation. Lastly, a novel set of virtual constraints was created that allowed us to realize experimentally the assumed CoM properties on a highly agile, 20 DoF bipedal robot. 

Currently, the performance of the robot is dependent on the ability of the operator to provide in real time an estimated terrain slope with respect to the body frame of the robot. Ideally, this information should be retrieved automously from a perception system, as in \cite{WaveFieldAutonomy2021}. When the slope is estimated accurately, the robot is very stable while walking with lateral and longitudinal velocity on sloped ground. In future work, we plan to: (a) improve step-to-step smoothness by appending a rate limiter related term to the cost function; (b) look at further relaxing the assumptions on the low-dimensional model (e.g., zero dynamics) to allow nonlinear terms; and (c) integrate the controller with a reactive planner.

\section*{Acknowledgment}
\small{ 
Toyota Research Institute provided funds to support this work. Funding for J. Grizzle
was in part provided by NSF Award No.~2118818. The authors thank Jennifer Humanchuk, Margaret Eva Mungai, Elizabeth Olson, and Jianyang Tang for their help in experiments. A portion of the experiments in the paper were conducted on Cassie Maize, thanks to a loan by the ROAHM Lab. 
}

\bibliographystyle{unsrt}
\balance
\bibliography{BibFiles/Bib2020July.bib,BibFiles/MPC.bib}
\balance
\end{document}